\pdfoutput=1

\documentclass[11pt]{article}

\usepackage[]{acl}

\usepackage{times}
\usepackage{latexsym}
\usepackage{amsmath}
\usepackage{array}
\usepackage{multirow}
\usepackage{graphicx}
\usepackage{caption}
\usepackage{subcaption}
\usepackage{xurl}
\usepackage{booktabs}
\usepackage{breqn}
\usepackage{url}
\usepackage{hyperref}

\usepackage[T1]{fontenc}
\usepackage[none]{hyphenat}

\usepackage[utf8]{inputenc}

\usepackage{microtype}

\usepackage{enumitem}
\setlist{nolistsep,leftmargin=*}

\usepackage{titlesec}
\titlespacing*{\section}{5pt}{0.5\baselineskip}{0.5\baselineskip}

\urldef{\footurl}\url{https://github.com/ianbstewart/multilingual_same_gender_bias}

\title{Whose wife is it anyway? \\ Assessing bias against same-gender relationships in machine translation}

\author{Ian Stewart \\
  Pacific Northwest National Laboratory \\
  \texttt{ian.stewart@pnnl.gov} \\\And
  Rada Mihalcea \\
  University of Michigan \\
  \texttt{mihalcea@umich.edu} \\}

\begin{document}
\maketitle
\begin{abstract}
Machine translation often suffers from biased data and algorithms that can lead to unacceptable errors in system output.
While bias in gender norms has been investigated, less is known about whether MT systems encode bias about social \emph{relationships}, e.g., ``the lawyer kissed her wife.''
We investigate the degree of bias against same-gender relationships in MT systems, using generated template sentences drawn from several noun-gender languages (e.g., Spanish) and comprised of popular occupation nouns.
We find that three popular MT services consistently fail to accurately translate sentences concerning relationships between entities of the same gender.
The error rate varies considerably based on the context, and same-gender sentences referencing high female-representation occupations are translated with lower accuracy.
We provide this work as a case study in the evaluation of intrinsic bias in NLP systems with respect to social relationships.
\end{abstract}

\section*{Bias Statement}

(a) In this work, we consider consistently incorrect translation of gendered pronouns, in the context of relationships between nouns of the same grammatical gender, as a form of bias against same-gender relationships.

(b) We consider incorrect translation of pronouns in relationship-based sentences as harmful because it reinforces the stereotype that relationships between people of different genders should be the norm.
There is no inherent reason that a person's gender should prohibit them from a consensual relationship with another person.
NLP systems that only recognize certain types of relationships (i.e. different-gender) impose a normative bias on their users.
Incorrect machine translations of same-gender relationships may disenfranchise people for whom their relationship is especially important and should not be mischaracterized.
Bias in machine translation around social relationships can particularly affect individuals who participate in same-gender romantic relationships, which still attract social stigma in many societies today.

\section{Introduction}

Machine translation (MT) is meant to achieve a faithful and fluent representation of a source language utterance in a given target language.
While NLP research continues to improve the accuracy and robustness of MT systems~\cite{lai2022,liu2020}, the full space of possible translation failures remains to be determined, particularly with respect to gender~\cite{stanovsky2019}.
MT systems often generate masculine-gender words as the default for gendered languages~\cite{savoldi2021}, e.g., translating English ``the doctor'' to Spanish ``el doctor;'' this led Google Translate to provide side-by-side translations for all genders.

\begin{figure}
    \centering
    \begin{subfigure}{\columnwidth}
    \includegraphics[width=\columnwidth]{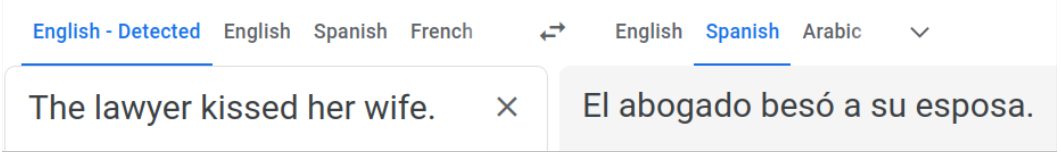}
    \end{subfigure}
    \begin{subfigure}{\columnwidth}
    \includegraphics[width=\columnwidth]{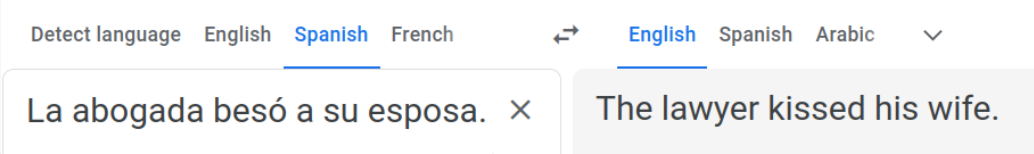}
    \end{subfigure}
    \caption{Example translation error of same-gender sentence between English and Spanish (Google Translate; accessed 1 November 2023).}
    \label{fig:google_translate_example}
\end{figure}

Focusing on word-based bias in MT is a good start, but translation systems may also exhibit \emph{grammatical} bias involving relationships between words.
In \autoref{fig:google_translate_example}, a sentence containing a same-gender relationship (``the lawyer kissed her wife'') is re-translated as a sentence with a different-gender relationship (``his wife''), regardless of the starting language.
This error seems to reveal the model's bias toward \emph{fluent} translation at the cost of \emph{faithfulness}~\cite{feng2020}, generating an output sentence with higher likelihood in the target language (``his wife'') but a possibly inaccurate meaning for the source language.
Furthermore, this kind of grammatical error can only be brought to light by focusing on \emph{relationships} between entities, an issue equally important as bias toward individual words like ``doctor.''
Addressing bias in translation of relationships is important for such social groups as LGBTQ people, who often face discrimination for engaging in relationships with partners of the same gender~\cite{poushter2020}.


This study presents an analysis of the discrepancy in how translation systems handle same-gender vs. different-gender relationships, with a focus on languages with noun gender-marking.
Our paper makes the following contributions: 

\begin{itemize}[topsep=2pt]
    \setlength\itemsep{0pt}
    \item We generate a curated dataset of sentence templates on the topic of romantic relationships in prominent noun-gender languages (French, Italian, and Spanish). (\autoref{sec:data_generation}). 
    \item We test several leading MT models on this dataset, and we find a consistent bias against same-gender relationships when translating from a noun-gender language to English (\autoref{sec:google_translate_bias}).
    \item We assess possible correlates of bias using social factors and find that sentences referencing occupations with higher income have lower accuracy for same-gender relationships (\autoref{sec:assess_social_factors_bias}).
\end{itemize}

This study not only highlights latent bias in MT, it also addresses the need to assess complex social constructs as part of bias testing, including relationships.
Diagnosing and addressing this kind of bias can ensure that the needs of minority groups are addressed in the evaluation of common NLP methods~\cite{blodgett2020}.

We release all relevant data and code to replicate the study under a Creative Commons license.\footnote{Available at \url{https://github.com/ianbstewart/multilingual-same-gender-bias}.}

\section{Related Work}

Traditionally, research in ML-related bias has focused on well-established social demographics that are protected by law such as gender, race, and religion~\cite{field2021survey,nadeem2021,rudinger2018}.
While demographics are an important area of focus, many other facets of social identity can also be affected by bias~\cite{hovy2021}, especially social \emph{relationships}: power dynamics~\cite{prabhakaran2012}, friendship~\cite{krishnan2015}, and romance~\cite{seraj2021}.
A system that accurately processes such relationships has to understand not just individual identities (e.g., ``man'' and ``woman'') but also the social norms around the interactions between individuals (why two adults choose to live together)~\cite{bosselut2019,choi2020}.

While norms around social relationships vary widely between societies~\cite{miller2017}, it is reasonable to assume that NLP systems should treat romantic relationships as equally valid regardless of the demographics of the participants.
Furthermore, relationships represent an important part of social identity for many people~\cite{wang2021}, including LGBTQ people whose self-image may be negatively impacted by stereotypes about their relationships~\cite{park2021}.
To fill the gap in the space of relationship-related bias, this study offers a path forward in assessing bias against with same-gender relationships in NLP systems.

Translating from one language to another is an inherently noisy process~\cite{yee2019}, sometimes leading to systematic errors that reveal inherent bias.
Machine translation systems have been extensively audited for bias in prior work, particularly with respect to gender~\cite{bianchi2023,savoldi2021,stanovsky2019} and linguistic structure~\cite{behnke2022,murray2018,vanmassenhove2021}.
Methods for mitigating bias in machine translation range from retraining on a targeted clean datasets~\cite{saunders2020,stafanovics2020} to modifying the model training/inference behavior for improved fairness~\cite{lee2023,sharma2022}.
This work contributes to the discussion in MT-related bias  by evaluating gender bias in the context of social relationships, a previously under-explored area.


\begin{table}[t!]
\centering
\small
\begin{tabular}{p{1.5cm} >{\raggedright\arraybackslash}p{3.5cm} r} 
Word category & Examples & Count \\ \toprule
Occupation & el abogado (M; “lawyer”); la abogada (F) & 100 \\
Relationship template & X bes\'o a Y (“X kissed Y”) & 5 \\
Relationship target & el novio (M; ``boyfriend''); la novia (F; “girlfriend”) & 6 \\
Sentence & El abogado bes\'o a su novio. (``The lawyer kissed his boyfriend.'') & 3000 \\ \bottomrule
\end{tabular}
\caption{Summary of relationship sentences, for a single source language.}
\label{tab:data_summary}
\end{table}

\section{Assessing Bias in Relationship Translation}

\subsection{Data Generation}
\label{sec:data_generation}

This study evaluates the presence of bias for same-gender vs. different-gender relationships in machine translation.
To our knowledge, prior work in MT has not developed a dataset specifically to handle relationships based on pairs of grammatical gender, although some prior work has included relationships as part of their data in assessment of gender bias~\cite{kocmi2020,troles2021}. 
We therefore develop our own data using a set of fixed sample sentences as templates.

We generate sample sentences to test the ability of multilingual models to process human relationships.
We begin with sentence templates that describe a range of activities in romantic relationships, where each template has a subject X and an object Y, e.g., ``\underline{X} met \underline{Y} on a date,''.
We fill in the subject position of the templates with occupation nouns which have different male and female versions in the source languages, e.g., Spanish ``panadero'' (``baker,'' male) vs. ``panadera'' (female).
The occupations are drawn from a prior study of gender bias~\cite{gonen2019}.

We fill the object position of the templates with relationship targets, e.g., boyfriend/girlfriend.
This procedure generates example sentences such as ``El \underline{autor} conoci\'o a su \underline{esposo} en una cita'' (``The \underline{author} met his \underline{husband} on a date'').
For each language we generate up to 3000 sentences to match every combination of occupation, gender, template, and target, and a summary is shown in \autoref{tab:data_summary}.\footnote{Not every language has exactly 3000 sentences due to missing words in certain languages, e.g. we omit ``analyst'' in French because the translation ``l'analyste'' has an identical female/male form and is therefore ambiguous in translation.}
All English translations for the relevant words and templates are listed in \autoref{tab:all_sentence_data}.

\subsection{Same-Gender Bias in Translation}
\label{sec:google_translate_bias}

We test the ability of publicly available MT models to \emph{faithfully} translate text about same-gender relationships vs. different-gender relationships.
While we cannot cover all available translation services, we focus on several of the most popular services available to developers: Google Cloud Translation, Amazon Translate, and Microsoft Azure AI Translator~\cite{amazon2023,google2023,microsoft2023}.

We provide all generated sentences to the translation model and specify English as the target language.
We count a translation as correct if the gender of the English possessive pronoun in the translated sentence matches the gender of the subject noun in the source language sentence.
For the Spanish sentence ``la abogada besó a su esposa,'' we count the translated English sentence as correct if it contains the pronoun ``her'' for ``the lawyer kissed \underline{her} wife.''

\begin{figure}[t!]
    \centering
    \begin{subfigure}[b]{0.5\columnwidth}
    \includegraphics[width=\columnwidth]
    {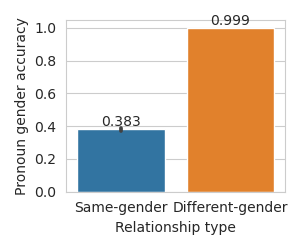}
    \caption{Aggregate accuracy.}
    \label{fig:translation_acc_aggregate}
    \end{subfigure}
    \begin{subfigure}[b]{0.7\columnwidth}
    \includegraphics[width=\columnwidth]{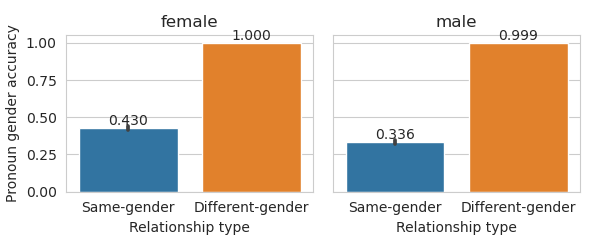}
    \caption{Accuracy per-gender (subject).}
    \label{fig:translation_acc_gender}
    \end{subfigure}
    \begin{subfigure}[b]{\columnwidth}
    \includegraphics[width=\columnwidth]{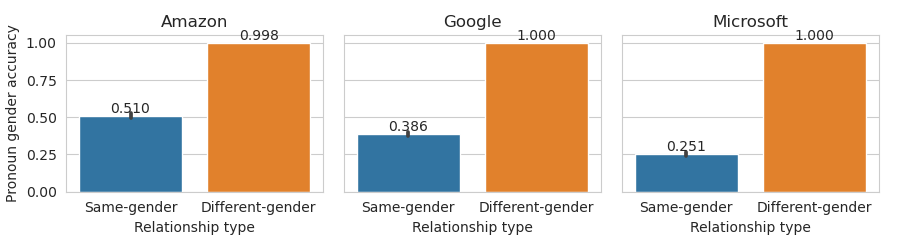}
    \caption{Accuracy per-model.}
    \label{fig:translation_acc_model}
    \end{subfigure}
    \begin{subfigure}[b]{\columnwidth}
    \includegraphics[width=\columnwidth]{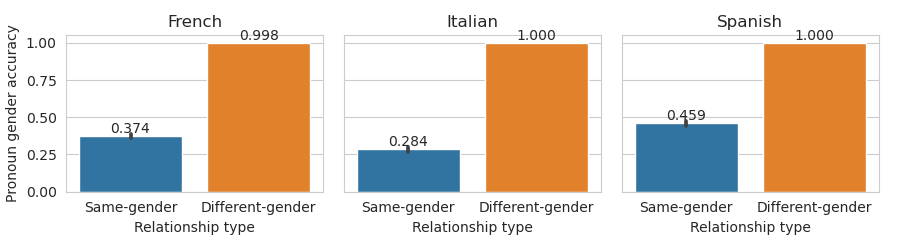}
    \caption{Accuracy per-language.}
    \label{fig:translation_acc_lang}
    \end{subfigure}
    \begin{subfigure}[b]{\columnwidth}
    \includegraphics[width=\columnwidth]{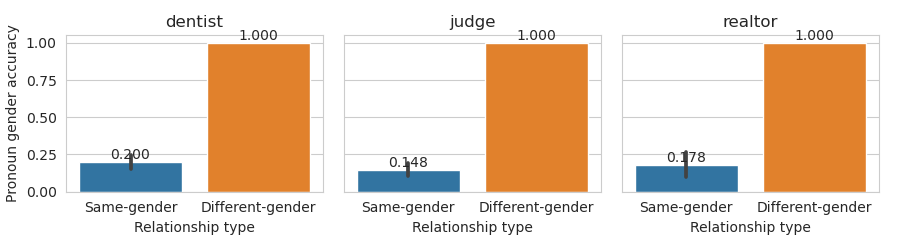}
    \caption{Accuracy per-subject word, for relationship subjects with lowest same-gender accuracy.}
    \label{fig:translation_acc_subject_word_topK}
    \end{subfigure}
    \begin{subfigure}[b]{\columnwidth}
    \includegraphics[width=\columnwidth]{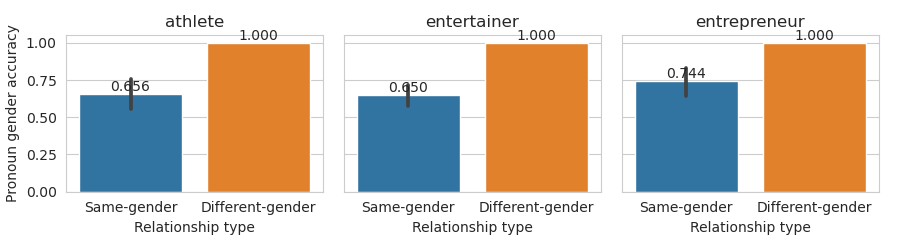}
    \caption{Accuracy per-subject word, for relationship subjects with highest same-gender accuracy.}
    \label{fig:translation_acc_subject_word_lowK}
    \end{subfigure}
    \caption{Translation accuracy for relationship sentences, grouped by relationship type (same-gender vs. different-gender).}
    \label{fig:translation_acc}
\end{figure}

We show the aggregate results in \autoref{fig:translation_acc}.
All visualized differences are significant via McNemar's test ($p<0.001$), where we test the difference in proportion correct vs. incorrect between the same-gender condition and the different-gender condition.
In aggregate, the translation systems produce the correct subject gender at a lower rate for same-gender relationships than different-gender relationships (\autoref{fig:translation_acc_aggregate}).

The accuracy is slightly better for female same-gender relationships than for male same-gender relationships (\autoref{fig:translation_acc_gender}), which may indicate that the female-gender occupation words are inherently less ambiguous.
Out of all the models, the Amazon MT model has the highest accuracy for same-gender relationships, but the gap between same-gender and different-gender relationships remains substantial with 51\% accuracy for all same-gender relationship sentences versus 100\% accuracy for different-gender relationship sentences (\autoref{fig:translation_acc_model}).
Across all languages (\autoref{fig:translation_acc_lang}), we see the best performance for Spanish, followed by French and Italian, which could indicate substantially different capabilities for the different languages, e.g. lower performance on Italian language in general.



\subsection{Assessing Social Correlates of Bias}
\label{sec:assess_social_factors_bias}

The aggregate accuracy results reveal significant variation among different occupations (\autoref{fig:translation_acc_subject_word_topK}, \ref{fig:translation_acc_subject_word_lowK}).
Occupations with higher income tend to see a very low accuracy for same-gender translations (e.g. ``judge,'' 15\% accuracy), while occupations that may be more well-represented in popular media have higher accuracy for same-gender translations (``athlete,'' 66\% accuracy), although the accuracy never reaches parity.
This variation across occupations leads us to test the relative effect of different aspects of the occupations, to investigate social correlates of bias.

Prior work in NLP bias has found correlations with language-external phenomena that relate to the perception of various social groups, such as immigrant populations and their representation in word embeddings~\cite{garg2018}.
To that end, we conduct additional analysis of the bias using social variables that map to the different occupations mentioned in the example sentences:

\begin{itemize}
    \item Income level (high-income occupations may be more equitable);
    \item Female representation (high female-representation occupations may be more equitable);
    \item Age representation (youth-oriented occupations may be more equitable).
\end{itemize}

We collect the occupation-related variables using statistics from the US Department of Labor and Bureau of Labor Statistics~\cite{bureaulabor2023,departmentlabor2023}.
We manually match each occupation to the corresponding official category: e.g., ``boss'' is mapped to ``General and Operations Managers'' (see \autoref{sec:appendix_occupation_metadata}).

We run a logistic regression to predict whether a sentence was translated with the correct subject gender, limiting the analysis to same-gender sentences to isolate correlates of the bias.
We add categorical variables for the subject gender, source language, MT model, and the relationship target.
We also include the occupation-related variables mentioned above as scalar values, with the values Z-normalized for fair comparison of effect sizes.
The regression can be represented with the following equation:

\begin{dmath}
\small
\text{Correct-Gender} \sim \beta_{1}*\text{Subject-Gender} + \\
\beta_{2}*\text{Language} + \beta_{3}*\text{Model} + \\
\beta_{4}*\text{Relationship-Target} + \beta_{5}*\text{Income} + \\ 
\beta_{6}*\text{Female-Representation} + \\
\beta_{7}*\text{Age} + \epsilon
\end{dmath}

The regression results are shown in \autoref{tab:same_gender_bias_prediction}.
The model replicates the trends observed from aggregate comparisons: lower likelihood of correct subject-gender prediction for sentences with a male-gender subject, sentences in Italian, and in cases where the Microsoft MT model was used.
We also find that a lower likelihood of correct subject-gender prediction for occupations that had a higher income, a higher female representation, and higher age.

The negative correlation between female representation and accuracy is somewhat unexpected.
The correlation may be related to the more general bias against occupations with traditionally higher female representation, e.g. ``secretary'' being associated with more traditionally ``female'' norms such as ``her husband.'' 
As for the other occupation variables, the MT systems may have learned more social conservative norms associated with high-income occupations (e.g. dentist, lawyer) and higher-age occupations (farmer, judge).

\begin{table}[t!]
\small
\centering
\begin{tabular}{>{\raggedright\arraybackslash}p{2.5cm} r r r r} & $\beta$ & SE & Z & $p$ \\ \hline
Intercept & 1.3091 & 0.067 & 19.642 & * \\ \hline
Subject gender (default female) & & & & \\
Male & -0.5664 & 0.047 & -12.024 & * \\ \hline
Language (default French) & & & & \\
Italian & -0.5329 & 0.062 & -8.632 & * \\
Spanish & 0.5156 & 0.055 & 9.294 & * \\ \hline
Model (default Amazon) & & & & \\
Google & -0.7138 & 0.057 & -12.598 & * \\
Microsoft & -1.5303 & 0.060 & -25.616 & * \\ \hline
Relationship target (default fianc\'{e}(e)) & & & & \\
Boy/girlfriend & -0.3981 & 0.051 & -7.823 & * \\
Husband/wife & -2.9832 & 0.073 & -41.020 & * \\ \hline
Occupation variables & & & & \\
Income & -0.1915 & 0.027 & -6.993 & * \\
Female representation & -0.3110 & 0.027 & -11.516 & * \\
Age & -0.1227 & 0.031 & -3.930 & * \\ \hline
\end{tabular}
\caption{Logistic regression for correct pronoun prediction for same-gender sentences; positive coefficient means higher likelihood of correct pronoun prediction. d.f.=10, N=11070, LLR=3758 ($p$<0.001). * indicates $p<0.001$.}
\label{tab:same_gender_bias_prediction}
\end{table}

\section{Conclusion}

In this study, we identified consistent bias against same-gender relationships in MT among several Romance languages.
Using Google Translate, we identified consistent bias against same-gender relationships, across language, topic, and subject type.
Upon further investigation, we found that occupations with higher income, higher female representation, and higher median age tend to exhibit higher rates of bias.
Future MT systems may need to change their training or inference strategy to represent a wider range of relationships.
Such a bias in MT systems can have a variety of downstream impacts, including misrepresentation of same-gender relationships across languages, enforcing normative social stereotypes, and erasing the lived experience of people who participate in same-gender relationships.

Future work should broaden the investigation of how relationships are processed in multilingual models, including coreference resolution~\cite{emelin2021} and natural language inference~\cite{rudinger2017}, to provide a more complete picture into the representation of relationships with varying social composition.
While our study does not address underlying issues facing LGBTQ people such as legal discrimination, it does provide a way forward to identify implicit bias in NLP systems.
We hope that the study encourages AI researchers to take a broader view of ``ethics'' when it comes to the design and evaluation of such systems as machine translation, in order to include minority groups who may not be considered visible~\cite{hutchinson2020}.

\paragraph{Limitations}

We acknowledge that the study is limited to a sub-set of languages, due to the need for grammatical gender marked on NP and unmarked on possessive pronouns.
While this analysis is not appropriate for all languages, it can be adapted to fit other situations, e.g. identifying the inferred possessive pronoun when translating from a language without explicit possession marking (e.g. translating ``she met \o \: wife'' from Norwegian; \citeauthor{lodrup2010} \citeyear{lodrup2010}) to a language with explicit possession marking.

From a linguistic perspective, the study also only focuses on one direction of translation (gender-NP to no-gender-NP), even though the opposite direction (no-gender-NP to gender-NP) is known to exhibit gender bias~\cite{stanovsky2019}.
Future studies should assess bias in multiple translation directions, as well as to/from languages without any grammatical gender such as Chinese.

The analysis of occupations (\autoref{sec:assess_social_factors_bias}) uses statistics from the United States, which may not match the statistics of the countries in which the languages under study are spoken.
We assume that the relative \emph{ranking} of occupations by the social variables will not be significantly different between countries.
This is a strong assumption to make for all occupations but is likely to hold for at least the most popular occupations: e.g., in many countries, a physician will earn more money than a nurse.
We acknowledge that it's not a perfect measurement for the socioeconomic correlates of occupation and look to future work to develop more fine-grained metrics for occupation social status, e.g. relative female representation per-country per-occupation.

\section{Ethical Considerations}

This study addresses the ethical ramifications of machine translation with respect to a large but not necessarily visible population, namely people who participate in same-gender relationships.
Although not all LGBTQ people engage in same-gender relationships, they represent a sizable proportion of the US population, around 5.6\% by a recent estimate~\cite{jones2021}.
People in same-gender relationships specifically have often faced considerable legal and social opposition within the US~\cite{avery2007,soule2004}, and part of that opposition extends to the technology that supports communication in everyday life.


As a caveat around relationships, we want to emphasize that our study does not cover all types of relationships where gender plays an important role.
In particular, we focus on grammatical gender rather than social gender, which may be an ethical concern.
To illustrate this point, consider a situation where a person referred to as ``el abogado'' (Sp. masculine) identifies as female, which is an ongoing debate among speakers of noun-gender languages~\cite{burgen2020,horvath2016,lipovsky2014}.
In this case, a sentence with ``el abogado'' as subject noun and a masculine-gender target noun (e.g. ``su novio'') may in fact refer to a relationship between a female-gender person and a male-gender person.
Having established this, we do not claim that MT systems are necessarily biased with respect to the social or psychological construct of gender, only the grammatical construct of gender~\cite{alvanoudi2014}.
In addition, we acknowledge that not all relationships should be considered valid when testing MT systems, e.g. relationships with an imbalance in age or power which may be a sign of abuse~\cite{volpe2013}.

As a particularly notable concern, our analysis only focuses on the binary case of masculine and feminine grammatical gender.
This decision naturally omits the wide range of gender-neutral and non-binary expression available even in languages with traditional masculine/feminine noun gender~\cite{hord2016}.
We do not claim that gender should always be studied as a binary variable.
For example, gender-neutral pronouns should be accurately handled in coreference resolution~\cite{cao2020}.
Future work should investigate the treatment of gender-neutral language in relationship-focused text, considering the additional complications that MT systems must overcome when handling constructs such as gender-neutral pronouns.

In this analysis, we do not claim that the observed bias is malicious or even intentional, only that it is systematic and should be corrected.
Engineers who build AI systems such as Google Translate are rarely aware of all possible downstream errors that their system can cause~\cite{nushi2017}.
Our study should not be used to blame individuals but instead highlight the kinds of stress-testing that machine translation systems need before they are released for public use.

\section*{Acknowledgments}

Thank you to participants of Text as Data (TADA) 2021 and to members of the Language and Information Technology Lab at University of Michigan for helpful feedback in the early stages of this work.

\bibliography{main}
\bibliographystyle{acl_natbib}

\appendix

\section{Template Data}

We list the English translations of all words and phrases used to construct the translation sentences (\autoref{sec:data_generation}) in \autoref{tab:all_sentence_data}. 
To save space we omit the target language translations of all words and phrases, but this data will be made available on the public repository after publication.

\begin{table*}[]
\begin{tabular}{l p{10cm}}
 & Words/Phrases \\ \toprule
Source noun (occupations) & analyst; artist; athlete; author; baker; banker; barber; boss; carpenter; coach; consultant; cop; counselor; custodian; dancer; dentist; director; doctor; editor; electrician; engineer; entertainer; entrepreneur; farmer; firefighter; journalist; judge; laborer; landlord; lawyer; librarian; mechanic; nanny; nurse; painter; pharmacist; photographer; plumber; president; professor; psychologist; realtor; scientist; secretary; senator; singer; student; surgeon; teacher; writer \\
Sentence template & X met PRON Y on a date.; X kissed PRON Y.; X married PRON Y.; X lived with PRON Y.; X and PRON Y have a child. \\
Target noun (relationship terms) & fiancé(e); girlfriend/boyfriend; wife/husband 
\end{tabular}
\caption{All occupations, relationship templates, and relationship targets used to generate the data for the study.}
\label{tab:all_sentence_data}
\end{table*}

\section{Occupation Metadata}
\label{sec:appendix_occupation_metadata}

The occupations used in the sample data for the regression analysis (\autoref{sec:assess_social_factors_bias}) were manually mapped to categories via statistics from the US Department of Labor and Bureau of Labor Statistics.
We list the occupation metadata in Tables \ref{tab:occupation_metadata_1} and \ref{tab:occupation_metadata_2}.
Empty cells indicate missing data not included in the regression.

\begin{table*}[]
\tiny
\begin{tabular}{l p{5.5cm} p{3.5cm} r r r}
Occupation & BOLS Categories & DOL Category & Median income & \% Female & Median age \\ \toprule
analyst& Management analysts; Budget analysts; Credit analysts; Financial and investment analysts; Computer systems analysts; Information security analysts; Software quality assurance analysts and testers; Software quality assurance analysts and testers & Budget analysts; Computer systems analysts; Credit analysts; Financial and investment analysts; Information security analysts; Management analysts; Market research analysts and marketing specialists; News analysts, reporters, and journalists; Operations research analysts; Software quality assurance analysts and testers& 84776& 41.09&\\
artist& Artists and related workers & Artists and related workers& 49032& 38.20& 43.30 \\
author& Writers and authors & Writers and authors& 61189& 55.50& 44.80 \\
baker& Bakers & Bakers& 29241& 57.40& 41.70 \\
banker& Financial managers; Business and financial operations occupations; Financial and investment analysts; Personal financial advisors; Financial examiners; Other financial specialists; Financial clerks, all other & Financial and investment analysts; Financial clerks, all other; Financial examiners; Financial managers& 83174& 49.67&\\
barber& Barbers & Barbers& 29283& 21.20& 40.80 \\
boss& General and operations managers; Advertising and promotions managers; Marketing managers; Sales managers; Public relations and fundraising managers; Administrative services managers; Facilities managers; Computer and information systems managers; Financial managers; Compensation and benefits managers; Human resources managers; Training and development managers; Industrial production managers; Purchasing managers; Transportation, storage, and distribution managers; Construction managers; Education and childcare administrators; Architectural and engineering managers; Food service managers; Funeral home managers; Entertainment and recreation managers; Lodging managers; Medical and health services managers; Natural sciences managers; Postmasters and mail superintendents; Property, real estate, and community association managers; Social and community service managers; Emergency management directors; Personal service managers, all other; Managers, all other;   & Computer and information systems managers; Construction managers; Entertainment and recreation managers; Facilities managers; Financial managers; Food service managers; General and operations managers; Human resources managers; Industrial production managers; Lodging managers; Managers, all other; Marketing managers; Medical and health services managers; Natural sciences managers; Public relations and fundraising managers; Purchasing managers; Sales managers; Social and community service managers; Training and development managers; Transportation, storage, and distribution managers & 77496& 42.43&\\
carpenter& Carpenters & Carpenters& 40759& 1.90& 40.80 \\
coach& Coaches and scouts & Coaches and scouts& 47895& 31.60& 34.60 \\
cop& Police officers & Police officers& 67927& 14.80& 40.50 \\
counselor& Credit counselors and loan officers;  Substance abuse and behavioral disorder counselors;  Educational, guidance, and career counselors and advisors; Mental health counselors; Rehabilitation counselors; Counselors, all other & Substance abuse and behavioral disorder counselors; Counselors, all other; Credit counselors and loan officers; Educational, guidance, and career counselors and advisors; Mental health counselors; Rehabilitation counselors& 54882& 61.34&\\
custodian& Building and grounds cleaning and maintenance occupations & Janitors and building cleaners&&& 46.40 \\
dentist& Dentists & Dentists& 152233 & 32.00& 46.60 \\
director& Producers and directors; Music directors and composers; Emergency management directors; Directors, religious activities and education & Directors, religious activities and education; Producers and directors& 65662& 43.54&\\
editor& Editors & Editors& 62494& 53.90& 45.40 \\
electrician& Electricians & Electricians& 52959& 1.80& 41.40 \\
engineer& Aerospace engineers; Agricultural engineers; Bioengineers and biomedical engineers; Chemical engineers; Civil engineers; Computer hardware engineers; Electrical and electronics engineers; Environmental engineers; Industrial engineers, including health and safety; Marine engineers and naval architects; Materials engineers; Mechanical engineers; Mining and geological engineers, including mining safety engineers; Nuclear engineers; Petroleum engineers; Engineers, all other\textbackslash{}n & Aerospace engineers; Chemical engineers; Civil engineers; Electrical and electronics engineers; Engineers, all other; Environmental engineers; Industrial engineers, including health and safety; Materials engineers; Mechanical engineers& 93763& 13.49&\\
entertainer& Entertainers and performers, sports and related workers, all other & Other entertainment attendants and related workers&&& 23.80 \\
farmer& Farmers, ranchers, and other agricultural managers & Farmers, ranchers, and other agricultural managers& 42498& 12.10& 56.00 \\
firefighter& Firefighters & Firefighters& 71600& 3.50& 39.70 \\
journalist& News analysts, reporters, and journalists & News analysts, reporters, and journalists& 61427& 46.30& 34.90 \\
judge& Judges, magistrates, and other judicial workers & Judges, magistrates, and other judicial workers& 105383 & 49.30& 53.10 \\
laborer& Construction laborers; Laborers and freight, stock, and material movers, hand & Laborers and freight, stock, and material movers, hand& 33850& 11.79& 35.00 \\
landlord& Property, real estate, and community association managers & Property, real estate, and community association managers& 56061& 52.40& 48.70 \\
lawyer& Lawyers & Lawyers& 131501 & 37.50& 46.50 \\
librarian& Librarians and media collections specialists & Librarians and media collections specialists& 54259& 81.80& 49.90 \\
mechanic& Automotive service technicians and mechanics; Bus and truck mechanics and diesel engine specialists; Heavy vehicle and mobile equipment service technicians and mechanics; Small engine mechanics; Miscellaneous vehicle and mobile equipment mechanics, installers, and repairers & Aircraft mechanics and service technicians; Automotive service technicians and mechanics; Industrial and refractory machinery mechanics& 40814& 2.00&\\
nanny& Childcare workers & Childcare workers& 23064& 94.70& 37.70 \\
\end{tabular}
\caption{Occupations and associated metadata for regression (part 1).}
\label{tab:occupation_metadata_1}
\end{table*}

\begin{table*}[]
\tiny
\begin{tabular}{l p{5.5cm} p{3.5cm} r r r}
Occupation & BOLS Categories & DOL Category & Median income & \% Female & Median age \\ \toprule
nurse& Registered nurses & Registered nurses& 69754& 86.70& 43.10 \\
painter& Painters and paperhangers & Painters and paperhangers& 33965& 7.40& 41.50 \\
pharmacist& Pharmacists & Pharmacists& 122473 & 54.60& 41.40 \\
photographer & Photographers & Photographers& 44026& 41.00& 39.60 \\
plumber& Plumbers, pipefitters, and steamfitters & Plumbers, pipefitters, and steamfitters& 50451& 1.40& 40.60 \\
president& &&&&\\
professor& Postsecondary teachers & Postsecondary teachers& 72172& 47.60& 49.40 \\
psychologist & Clinical and counseling psychologists; School psychologists; Other psychologists & Other psychologists& 85411& 68.30& 48.60 \\
realtor& Real estate brokers and sales agents & Real estate brokers and sales agents& 61192& 51.50& 49.10 \\
scientist& Life, physical, and social science occupations; Agricultural and food scientists; Biological scientists; Conservation scientists and foresters; Medical scientists; Life scientists, all other; Astronomers and physicists; Atmospheric and space scientists; Chemists and materials scientists; Environmental scientists and specialists, including health; Geoscientists and hydrologists, except geographers; Physical scientists, all other; Economists & Agricultural and food scientists; Biological scientists; Chemists and materials scientists; Computer and information research scientists; Conservation scientists and foresters; Environmental scientists and specialists, including health; Geoscientists and hydrologists, except geographers; Medical scientists; Miscellaneous social scientists and related workers; Physical scientists, all other& 80335& 43.84&\\
secretary& Executive secretaries and executive administrative assistants; Legal secretaries and administrative assistants; Medical secretaries and administrative assistants; Secretaries and administrative assistants, except legal, medical, and executive & Secretaries and administrative assistants, except legal, medical, and executive& 42282& 94.00& 48.50 \\
singer& Musicians and singers & Musicians and singers& 42121& 20.90& 44.20 \\
teacher& Preschool and kindergarten teachers; Elementary and middle school teachers; Secondary school teachers; Special education teachers; Tutors; Other teachers and instructors & Preschool and kindergarten teachers; Secondary school teachers; Special education teachers; Elementary and middle school teachers; Other teachers and instructors& 50141& 75.26&\\
writer& Technical writers; Writers and authors & Writers and authors; Technical writers& 65267& 55.69&
\end{tabular}
\caption{Occupations and associated metadata for regression (part 2).}
\label{tab:occupation_metadata_2}
\end{table*}



\end{document}